\documentclass[]{elsarticle}

\usepackage{graphicx}
\usepackage{amssymb}
\usepackage{bm}
\usepackage{xcolor}
\usepackage{amsmath}

\usepackage{hyperref}

\journal{Journal of Medical Physics}

\begin{document}

\begin{frontmatter}


\title{A Gentle Introduction to Deep Learning in Medical Image Processing}



\author{Andreas Maier$^1$, Christopher Syben$^1$, Tobias Lasser$^2$, Christian Riess$^1$}

\address{$^1$Friedrich-Alexander-University Erlangen-Nuremberg, Germany\\$^2$Technical University of Munich, Germany}

\begin{abstract}
This paper tries to give a gentle introduction to deep learning in medical image processing, proceeding from theoretical foundations to applications. We first discuss general reasons for the popularity of deep learning, including several major breakthroughs in computer science. Next, we start reviewing the fundamental basics of the perceptron and neural networks, along with some fundamental theory that is often omitted. Doing so allows us to understand the reasons for the  rise of deep learning in many application domains. Obviously medical image processing is one of these areas which has been largely affected by this rapid progress, in particular in image detection and recognition, image segmentation, image registration, and computer-aided diagnosis. There are also recent trends in physical simulation, modelling, and reconstruction that have led to astonishing results. Yet, some of these approaches neglect prior knowledge and hence bear the risk of producing implausible results.  These apparent weaknesses highlight current limitations of deep learning. However, we also briefly discuss promising approaches that might be able to resolve these problems in the future.
\end{abstract}


\end{frontmatter}

\newcommand{\comment}{\color{black}}
\newcommand{\commenttwo}{\color{black}}
\section{Introduction}
\label{S:1}

Over the recent years, Deep Learning (DL) \cite{lecun2015deep} has had a tremendous impact on various fields in science. It has lead to significant improvements in speech recognition \cite{dahl2012context} and image recognition \cite{krizhevsky2012imagenet}, it is able to train artificial agents that beat human players in Go \cite{silver2016mastering} and ATARI games \cite{mnih2015human}, and it creates artistic new images \cite{mordvintsev2015inceptionism, tan2017artgan} and music \cite{DBLP:journals/corr/abs-1709-01620}. 
Many of these tasks were considered to be impossible to be solved by computers before the advent of deep learning, even in science fiction literature.

Obviously this technology is also highly relevant for medical imaging. Various introductions to the topic can be found in the literature ranging from short tutorials and reviews \cite{seebock2015deep,  shen2017deep, pawlowski2017dltk, litjens2017survey, erickson2017machine, suzuki2017survey, hagerty2017medical, lakhani2018hello, kim2018prospects, ker2018deep} over blog posts and jupyter notebooks \cite{rajchl2018introduction, breininger2018tutorial, cornelisse2018} to entire books \cite{zhou2017deep, lu2017deep, chollet2017deep, geron2017hands}. All of them serve a different purpose and offer a different view on this quickly evolving topic. A very good review paper is for example found in the work of Litjens et al. \cite{litjens2017survey}, as they did the incredible effort to review more than 300 papers in their article. Since then, however, many more noteworthy works have appeared - almost on a daily basis - which makes it difficult to create a review paper that matches the current pace in the field. {\comment The newest effort to summarize the entire field was attempted in \cite{sahiner2018deep} listing more than 350 papers. Again, since its publication several more noteworthy works appeared and others were missed.} Hence, it is important to select methods of significance and describe them in high detail. Zhou et al. \cite{zhou2017deep} do so for the state-of-the-art of deep learning in medical image analysis and found an excellent selection of topics. Still, deep learning is being quickly adopted in other fields of medical image processing and the book misses, for example, topics such as image reconstruction. While an overview on important methods in the field is crucial, the actual implementation is as important to move the field ahead. Hence, works like the short tutorial by Breininger et al. \cite{breininger2018tutorial} are highly relevant to introduce to the topic also on a code-level. Their jupyter notebook framework creates an interactive experience in the web browser to implement fundamental deep learning basics in Python. In summary, we observe that the topic is too complex and evolves too quickly to be summarized in a single document. Yet, over the past few months there already have been so many exciting developments in the field of medical image processing that we believe it is worthwhile to point them out and to connect them to a single introduction.

Readers of this article do not have to be closely acquainted with deep learning at its terminology. We will summarize the relevant theory and present it at a level of detail that is sufficient to follow the major concepts in deep learning. Furthermore, we connect these observations with traditional concepts in pattern recognition and machine learning. In addition, we put these foundations into the context of emerging approaches in medical image processing and analysis, including applications in physical simulation and image reconstruction. As a last aim of this introduction, we also clearly indicate potential weaknesses of the current technology and outline potential remedies.

\section{Materials and Methods}
\newcommand{\fex}{\bm x}
\newcommand{\param}{\bm \theta}
\newcommand{\activation}{h}
\newcommand{\real}{{\rm I\!R}}
\newcommand{\prediction}{\hat{y}}
\newcommand{\classvar}{y}
\newcommand{\classifier}{\hat{f}}
\newcommand{\loss}{L}

\begin{figure}
    \centering
    \includegraphics[width=\linewidth]{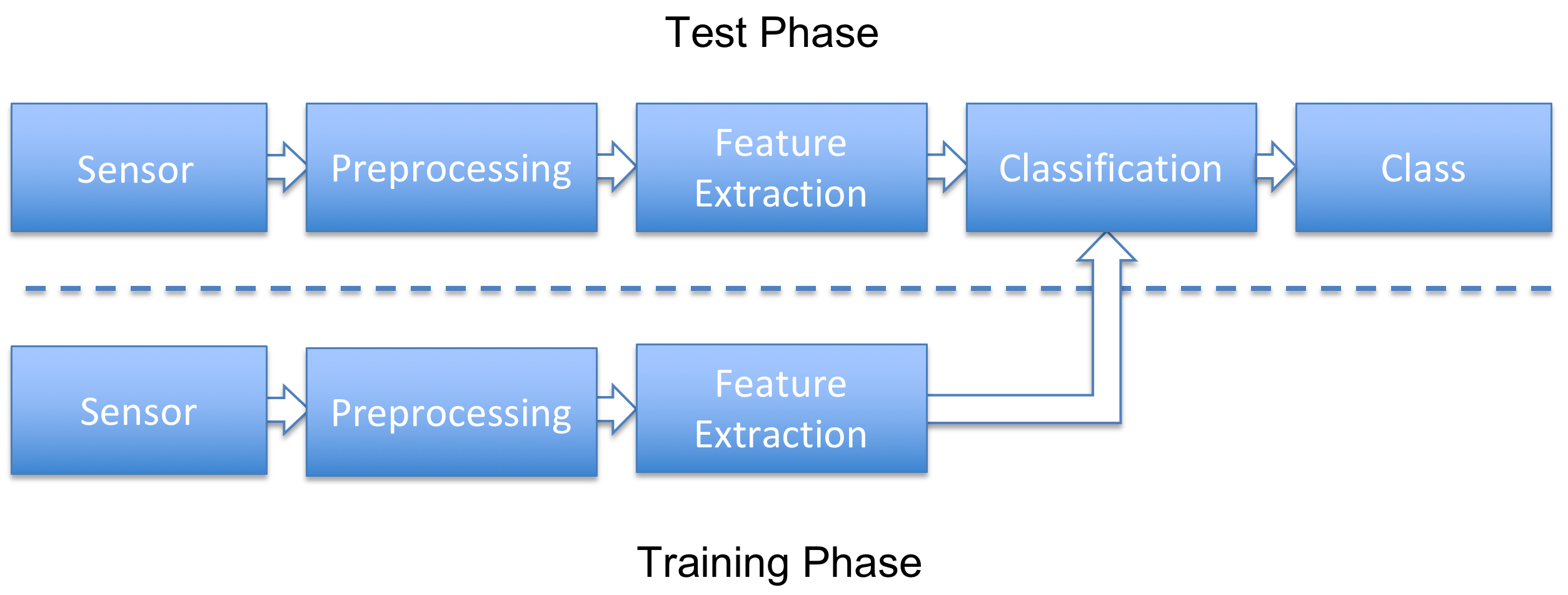}
    \caption{Schematic of the traditional pattern recognition pipeline used for automatic decision making. Sensor data is preprocessed and ``hand-crafted'' features are extracted in training and test phase. During training a classifier is trained that is later used in the test phase to decide the class automatically (after \cite{niemann2013pattern}).}
    \label{fig:pattern_rec_pipeline}
\end{figure}

\subsection{Introduction to machine learning and pattern recognition}

Machine learning and pattern recognition essentially deal with the problem of automatically finding a decision, for example, separating apples from pears. In traditional literature \cite{niemann2013pattern}, this process is outlined using the pattern recognition system (cf. Fig.~\ref{fig:pattern_rec_pipeline}). During a training phase, the so-called \emph{training data set} is \emph{preprocessed} and meaningful \emph{features} are extracted. While the preprocessing is understood to remain in the original space of the data and comprised operations such as noise reduction and image rectification, feature extraction is facing the task to determine an algorithm that would be able to extract a distinctive and complete feature representation, for example, color or length of the semi-axes of a surrounding ellipse for our apples and pears example. This task is truly difficult to generalize, and it is necessary to design such features anew essentially for every new application. In the deep learning literature, this process is often also referred to as ``hand-crafting'' features. Based on the feature vector $\fex \in \real^n$, the \emph{classifier} has to predict the correct \emph{class} $\classvar$, which is typically estimated by a function $\prediction = \classifier(\fex)$ that directly results in the classification result $\prediction$. The classifier's parameter vector $\param$ is determined during the training phase and later evaluated on an independent \emph{test data set}.

\subsection{Neural networks}
\begin{figure}
    \centering
    \includegraphics[width=\linewidth]{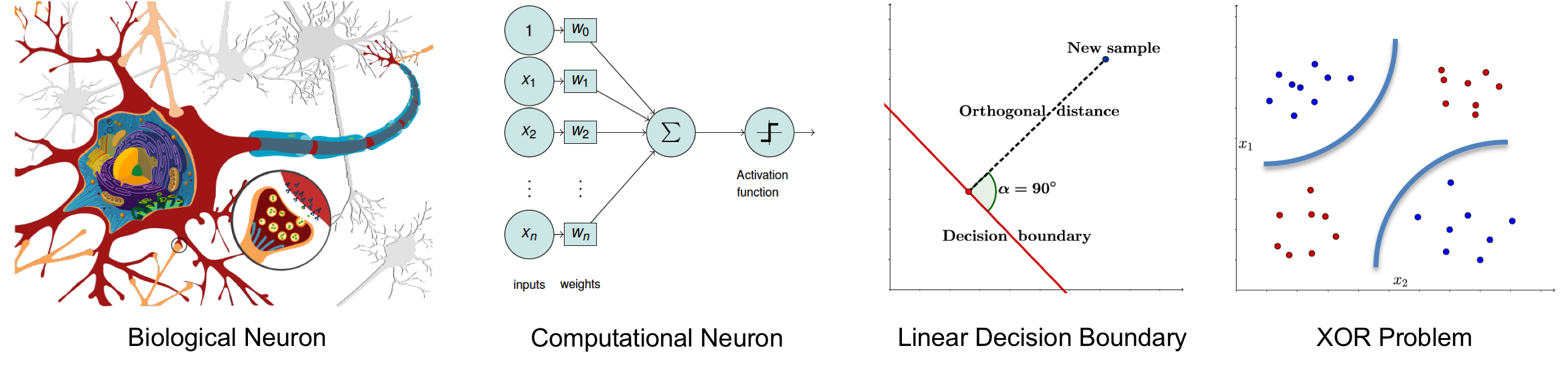}
    \caption{Neurons are inspired by biological neurons shown on the left. The resulting computational neuron computes a weighted sum of its inputs which is then processed by an activation function $\activation(x)$ to determine the output value (cf. Fig.~\ref{fig:activations}). Doing so, we are able to model linear decision boundaries, as the weighted sum can be interpreted as a signed distance to the decision boundary, while the activation determines the actual class membership. On the right-hand side, the XOR problem is shown that cannot be solved by a single linear classifier. It typically requires either curved boundaries or multiple lines.}
    \label{fig:perceptron}
\end{figure}

In this context, we can now follow neural networks and associated methods in their role as classifiers.
The fundamental unit of a neural network is a neuron, it takes a bias $w_0$ and a weight vector $\bm w=(w_1, \ldots, w_n)$ as parameters $\param = (w_0,\ldots,w_n)$ to model a decision
\begin{equation}
    \classifier(\fex) = \activation(\bm w^\top\fex + w_0)
\end{equation}
using a non-linear activation function $\activation(x)$. Hence, a single neuron itself can already be interpreted as a classifier, if the activation function is chosen such that it is monotonic, bounded, and continuous. In this case, the maximum and the minimum can be interpreted as a decision for the one or the other class. Typical representatives for such activation functions in classical literature are the sign function $\textrm{sign}(x)$ resulting in Rosenblatt's perceptron \cite{rosenblatt1957perceptron}, the sigmoid function $\sigma(x) = \frac{1}{1 + e^{-x}}$, or the tangens hyperbolicus $\textrm{tanh}(x)=\frac{e^x-e^{-x}}{e^x+e^{-x}}$. (cf. Fig.~\ref{fig:activations}). A major disadvantage of individual neurons is that they only allow to model linear decision boundaries, resulting in the well known fact that they are not able to solve the \emph{XOR} problem. Fig.~\ref{fig:perceptron} summarizes the considerations towards the computational neuron graphically.

In combination with other neurons, modelling capabilities increase dramatically. Arranged in a \emph{single layer}, it can already be shown that neural networks can approximate any continuous function $f(\fex)$ on a compact subset of $\real^n$ \cite{cybenko1989approximation}. A single layer network is conveniently summarized as a linear combination of $N$ individual neurons 
\begin{equation}
    \classifier(\fex) = \sum_{i=0}^{N-1} v_i \activation({\bm w}_i^\top \fex +w_{0,i})
\end{equation}
using combination weights $v_i$. All trainable parameters of this network can be summarized as 
$$\param = (v_0, w_{0,0}, {\bm w}_0, \ldots, v_N, w_{0,N}, {\bm w}_N)^\top.$$
The difference between the true function $f(\fex)$ and its approximation $\classifier(\fex)$ is bounded by 
\begin{equation}
    |f(\fex)-\classifier(\fex)| < \epsilon,
\end{equation}
where $\epsilon$ decreases with increasing $N$ for activation functions that satisfy the criteria that we mentioned earlier (monotonicity, boundedness, continuity) \cite{hornik1991approximation}. Hence, given a large number of neurons, \emph{any function can be approximated using a single layer network only}. {\comment Note that the approximation will only be valid for samples that are drawn from the same compact set on which the network was trained. As such, an additional practical requirement for an approximation is that the training set is \emph{representative} and future observations will be similar.} At first glance, this contradicts all recent developments in deep learning and therefore requires additional attention.

\begin{figure}[t]
    \centering
    \includegraphics[width=\linewidth]{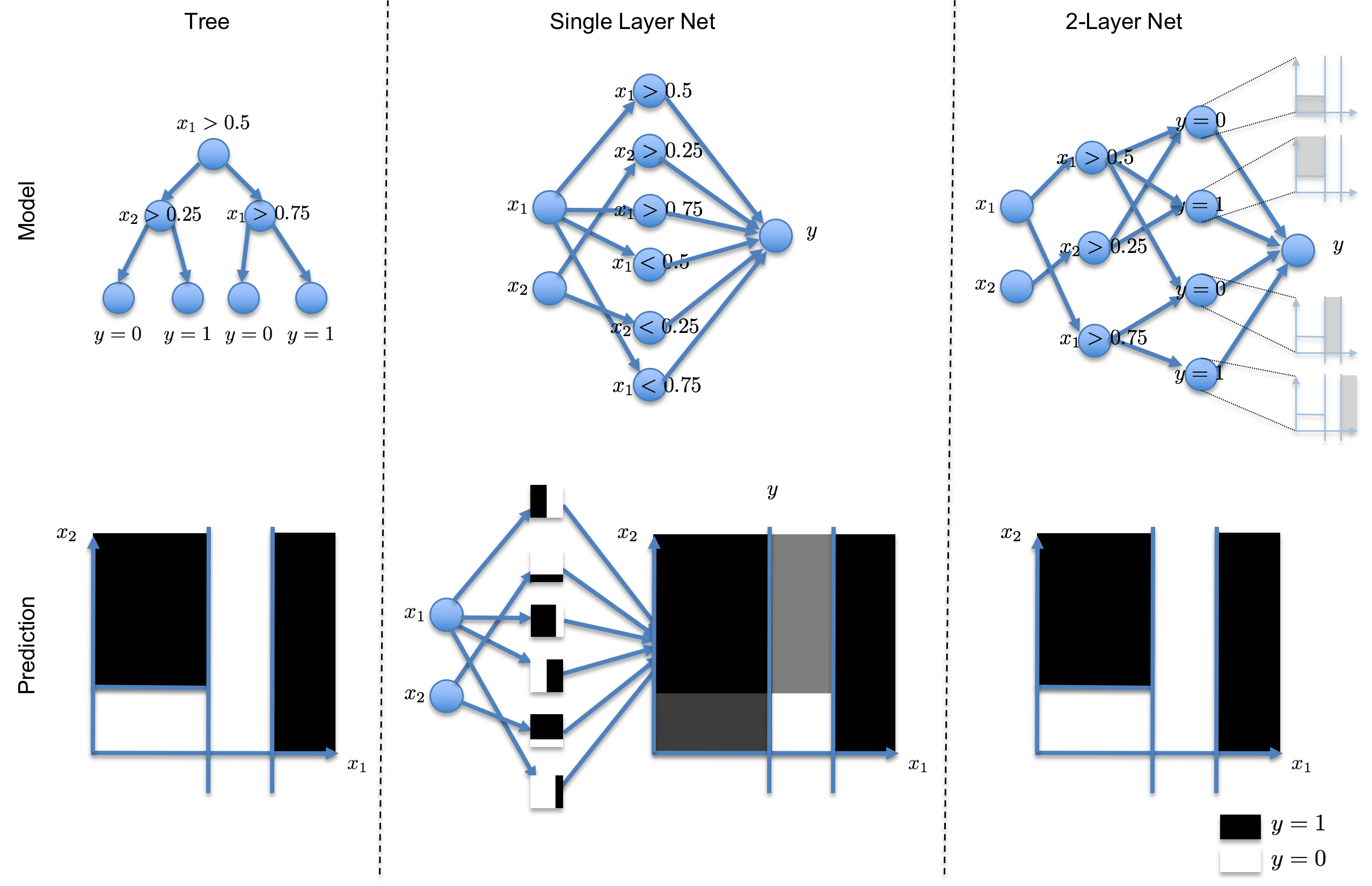}
    \caption{A decision tree allows to describe any partition of space and can thus model any decision boundary. Mapping the tree into a one-layer network is possible. Yet, there still is significant residual error in the resulting function. In the center example, $\epsilon \approx 0.7$. In order to reduce this error further, a higher number of neurons would be required. If we construct a network with one node for every inner node in the first layer and one node for every leaf node in the second layer, we are able to construct a network that results in $\epsilon = 0$.}
    \label{fig:treeexample}
\end{figure}

In the literature, many arguments are found why a deep structure has benefits for feature representation, including the argument that by recombination of the weights along the different paths through the network, features may be re-used exponentially \cite{bengio2013representation}. Instead of summarizing this long line of arguments, we look into a slightly simpler example that is summarized graphically in Fig.~\ref{fig:treeexample}. Decision trees are also able to describe general decision boundaries in $\real^n$. A simple example is shown on the top left of the figure, and the associated partition of a two-dimensional space is shown below, where black indicates class $y=1$ and white $y=0$. According to the universal approximation theorem, we should be able to map this function into a single layer network. In the center column, we attempt to do so using the inner nodes of the tree and their inverses to construct a six neuron basis. In the bottom of the column, we show the basis functions that are constructed at every node projected into the input space, and the resulting network's approximation, also shown in the input space. Here, we chose the output weights to minimize $||\classvar-\prediction||_2$. As can be seen in the result, not all areas can be recovered correctly. In fact, the maximal error $\epsilon$ is close to 0.7 for a function that is bounded by 0 and 1. In order to improve this approximation, we can choose to introduce a second layer. As shown in the right column, we can choose the strategy to map all inner nodes to a first layer and all leaf nodes of the tree to a second layer. Doing so effectively encodes every partition that is described by the respective leaf node in the second layer. This approach is able to map our tree correctly with $\epsilon = 0$. In fact, this approach is general, holds for all decision trees, and was already described by Ivanova et al. in 1995 \cite{ivanova1995initialization}. As such, we can now understand why deeper networks may have more modelling capacity.

\subsection{Network training}
\label{sec:training}
Having gained basic insights into neural networks and their basic topology, we still need to discuss how its parameters $\param$ are actually determined. The answer is fairly easy: gradient descent. In order to compute a gradient, we need to define a function that measures the quality of our parameter set $\param$, the so-called \emph{loss function} $\loss(\param)$. In the following, we will work with simple examples for loss functions to introduce the concept of \emph{back-propagation}, which is the algorithm that is commonly used to efficiently compute gradients for neural network training.

We can represent a single-layer fully connected network with linear activations simply as $\bm \prediction = \bm \classifier(\fex) = \bm{W} \fex$, i.e., a matrix multiplication. Note that the network's output is now multidimensional with $\bm \prediction, \bm \classvar \in \real^m$. Using an L2-loss, we end up with the following objective function:
\begin{equation}
    \loss(\param) = \frac{1}{2} ||\bm \classifier(\fex) - \bm \classvar ||^2_2 =  \frac{1}{2} || \bm{W} \fex - \bm \classvar ||^2_2.
\end{equation}
In order to update the parameters $\param = \bm W$ in this example, we need to compute
\begin{equation}
    \frac{\partial \loss}{\partial \bm W} = \underbrace{\frac{\partial \loss}{\partial \bm \classifier}}_{(\bm{W} \fex - \bm \classvar)} \underbrace{\frac{\partial \bm \classifier}{\partial \bm W}}_{\cdot(\fex^\top)} = (\bm{W} \fex - \bm \classvar) (\fex^\top)
\end{equation}
using the chain rule. Note that $\cdot$ indicates the operator's side, as matrix vector multiplications generally do not commute. The final weight update is then obtained as
\begin{equation}
    \bm{W}^{j+1}= \bm W^j + \eta (\bm{W}^j \fex - \bm \classvar) \fex^\top,
    \label{eq:update}
\end{equation}
where $\eta$ is the so-called \emph{learning rate} and $j$ is used to index the iteration number.

\begin{figure}
    \centering
    \includegraphics[width=\linewidth]{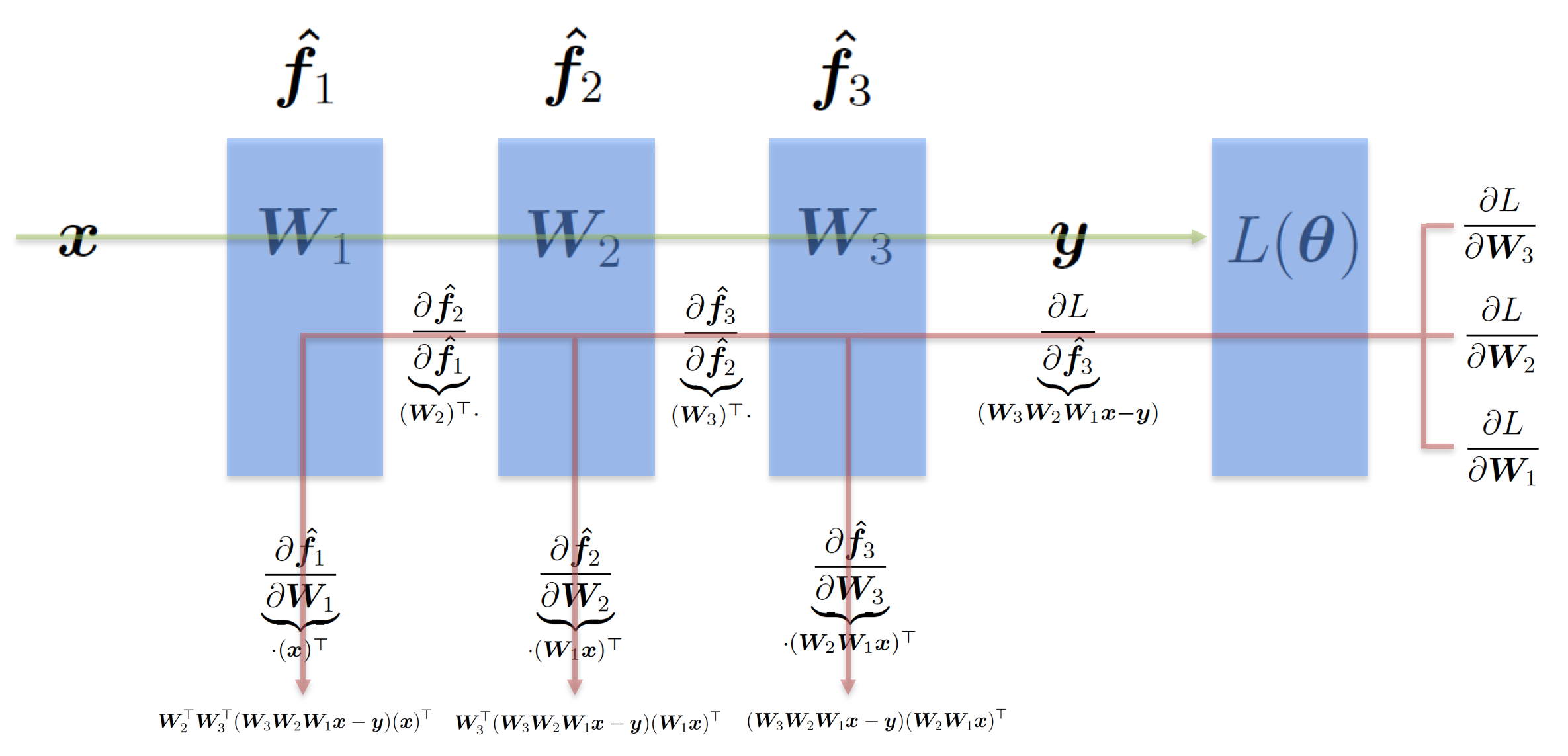}
    \caption{Graphical overview of back-propagation using layer derivatives. During the forward pass, the network is evaluated once and compared to the desired  output using the loss function. The back-propagation algorithm follows different paths through the layer graph in order to compute the matrix derivatives efficiently.}
    \label{fig:backprop}
\end{figure}

Now, let us consider a slightly more complicated network structure with three layers $\bm \prediction = \bm \classifier_3 (\bm \classifier_2 ( \bm \classifier_1 (\fex))) = \bm W_3 \bm W_2 \bm W_1 \fex$, again using linear activations. This yields the following objective function:
\begin{equation}
    \loss(\param) = \frac{1}{2} ||\bm W_3 \bm W_2 \bm W_1 \fex - \bm \classvar ||^2_2.
\end{equation}
Note that this example is academic, as $\param = \{\bm W_1, \bm W_2, \bm W_3\}$ could simply be collapsed to a single matrix. {\comment Yet, the concept that we use to derive this gradient is generally applicable also to non-linear functions.} Computing the gradient with respect to the parameters of the last layer $\bm W_3$ follows the same recipe as in the previous network:
\begin{equation}
    \frac{\partial \loss}{ \partial \bm W_3} = \underbrace{\frac{\partial \loss}{ \partial \bm \classifier_3}}_{(\bm W_3 \bm W_2 \bm W_1 \fex - \bm \classvar)} \underbrace{\frac{\partial \bm \classifier_3}{ \partial \bm W_3}}_{\cdot(\bm W_2 \bm W_1 \fex)^\top} = (\bm W_3 \bm W_2 \bm W_1 \fex - \bm \classvar) {(\bm W_2 \bm W_1 \fex)^\top}.
\end{equation}
For the computation of the gradient with respect to the second layer $\bm W_2$, we already need to apply the chain rule twice:
\begin{eqnarray}
    \frac{\partial \loss}{ \partial \bm W_2} &=& \frac{\partial \loss}{ \partial \bm \classifier_3} \frac{\partial \bm \classifier_3}{ \partial \bm W_2} = \underbrace{  \frac{\partial \loss}{ \partial \bm \classifier_3}}_{(\bm W_3 \bm W_2 \bm W_1 \fex - \bm \classvar)} \underbrace{\frac{\partial \bm \classifier_3}{ \partial \bm \classifier_2}}_{(\bm W_3)^\top\cdot} \underbrace{\frac{\partial \bm \classifier_2}{ \partial \bm W_2}}_{\cdot(\bm W_1 \fex)^\top}\nonumber
    \\
    &=&\bm W_3^\top {(\bm W_3 \bm W_2 \bm W_1 \fex - \bm \classvar)} {( \bm W_1 \fex)^\top}.
\end{eqnarray}
Which leads us to the input layer gradient that is determined as
\begin{eqnarray}
    \frac{\partial \loss}{ \partial \bm W_1} &=& \frac{\partial \loss}{ \partial \bm \classifier_3} \frac{\partial \bm \classifier_3}{ \partial \bm W_1} = 
     \frac{\partial \loss}{ \partial \bm \classifier_3} 
     \frac{\partial \bm \classifier_3}{ \partial \bm \classifier_2}
     \frac{\partial \bm \classifier_2}{ \partial \bm W_1}
    =
    \underbrace{  \frac{\partial \loss}{ \partial \bm \classifier_3}}_{(\bm W_3 \bm W_2 \bm W_1 \fex - \bm \classvar)} 
    \underbrace{\frac{\partial \bm \classifier_3}{ \partial \bm \classifier_2}}_{(\bm W_3)^\top\cdot} 
    \underbrace{\frac{\partial \bm \classifier_2}{ \partial \bm \classifier_1}}_{(\bm W_2)^\top\cdot}
    \underbrace{\frac{\partial \bm \classifier_1}{ \partial \bm W_1}}_{\cdot(\fex)^\top}\nonumber
    \\
    &=&\bm W_2^\top \bm W_3^\top {(\bm W_3 \bm W_2 \bm W_1 \fex - \bm \classvar)} {(\fex)^\top}. \label{eq:layer1}
\end{eqnarray}
The matrix derivatives above are also visualized graphically in Fig.~\ref{fig:backprop}. Note that many intermediate results can be reused during the computation of the gradient, which is one of the reasons why back-propagation is efficient in computing updates. Also note that the forward pass through the net is part of $\frac{\partial \loss}{ \partial \bm \classifier_3}$, which is contained in all gradients of the net. The other partial derivatives are only partial derivatives either with respect to the input or the parameters of the respective layer. Hence, back-propagation can be used if both operations are known for every layer in the net. Having determined the gradients, each parameter can now be updated analogous to Eq.~\ref{eq:update}.

\subsection{Deep learning}
\begin{figure}
    \centering
    \includegraphics[width=\linewidth]{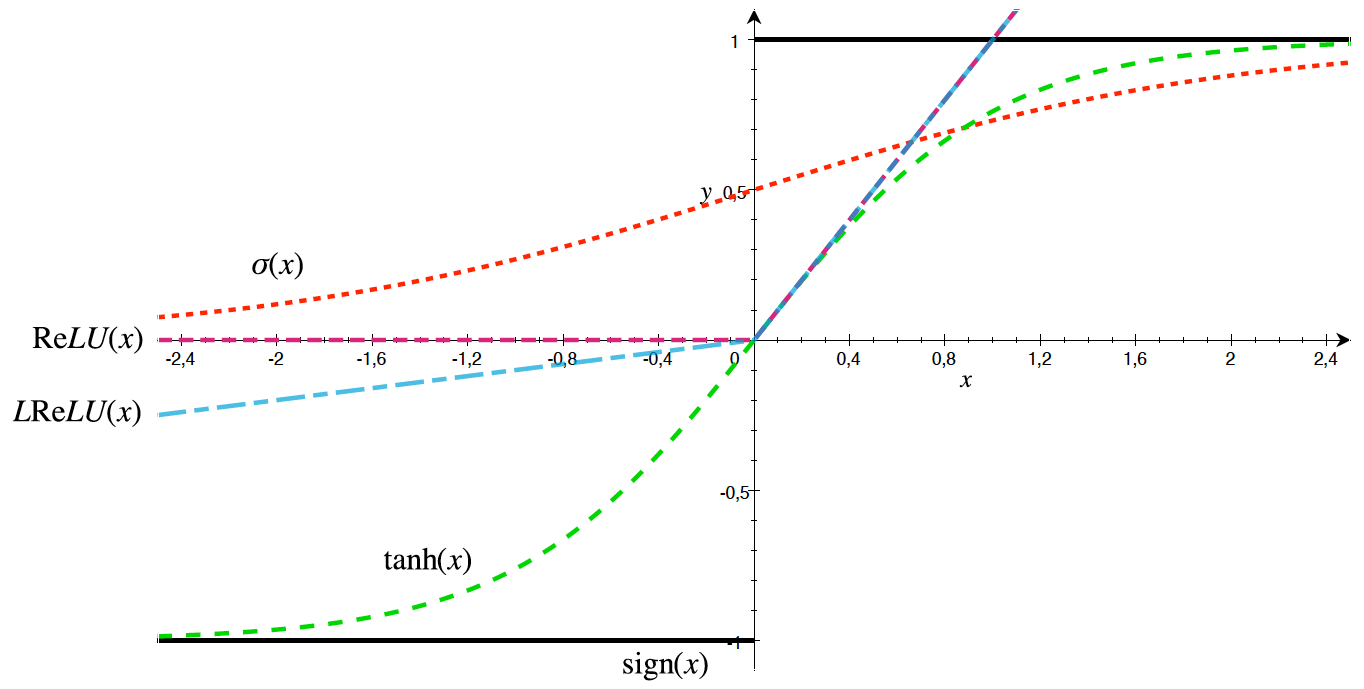}
    \caption{Overview of classical ($\textrm{sign}(x)$, $\sigma(x)$, and $\textrm{tanh}(x)$) and modern activation functions, like the Rectified Linear Unit $\textrm{ReLU}(x)$ and the leaky ReLU $\textrm{LReLU}(x)$.}
    \label{fig:activations}
\end{figure}

With the knowledge summarized in the previous sections, networks can be constructed and trained. However, deep learning is not possible. One important element was the establishment of additional activation functions that are displayed in Fig.~\ref{fig:activations}. In contrast to classical bounded activations like $\textrm{sign}(x)$, $\sigma(x)$, and $\textrm{tanh}(x)$, the new functions such as the \emph{Rectified Linear Unit} {\comment
$$\textrm{ReLU}(x) = \begin{cases} x & \textrm{if}~x \geq 0 \\ 0 & \textrm{else,}\end{cases}$$}
and many others, of which we only mention the \emph{Leaky ReLU}
$$\textrm{LReLU}(x) = \begin{cases} x & \textrm{if}~x \geq 0 \\ \alpha x & \textrm{else,}\end{cases}$$
were identified to be useful to train deeper networks. Contrary to the classical activation functions, many of the new activation functions are \emph{convex} and have large areas with non-zero derivatives.
{\comment As can be seen in Eq.~\ref{eq:layer1}, the computation of the gradient of deeper layers using the chain rule requires several multiplications of partial derivatives. The deeper the net, the more multiplications are required. If several elements along this chain are smaller than $1$, the entire gradient decays exponentially with the number of layers. Hence, non-saturating derivatives are important to
solve numerical issues, which were the reasons why \emph{vanishing gradients} did not allow training of networks that were much deeper than about three layers.}
Also note that each neuron does not loose its interpretation as a classifier, if we consider 0 as the classification boundary. Furthermore, the universal approximation theorem still holds for a single-layer network with ReLUs \cite{sonoda2017neural}. Hence, several useful and desirable properties are attained using such modern activation functions. 

One disadvantage is, of course, that the ReLU is not differentiable over the entire domain of $x$. At $x=0$ a kink is found that does not allow to determine a unique gradient. For optimization, an important property of the gradient of a function is that it will point towards the direction of the steepest ascent. Hence, following the negative direction will allow minimization of the function. For a differentiable function, this direction is unique. If this constraint is relaxed to allow multiple directions that lead to an extremum, we arrive at sub-gradient theory \cite{rockafellar}. It allows us to still use gradient descent algorithms to optimize such problems, if it is possible to determine a \emph{sub-gradient}, i.e., at least one instance of a valid direction towards the optimum. For the ReLU, any value between 0 and -1 would be acceptable at $x=0$ for the descent operation. If such a direction can be obtained, convergence is guaranteed for convex problems by application of specific optimization programs, such as using a fixed step size in the gradient descent \cite{bertsekas2015convex}. This allows us to remain with back-propagation for optimization, while using non-differentiable activation functions.

\begin{figure}
    \centering
    \includegraphics[width=\linewidth]{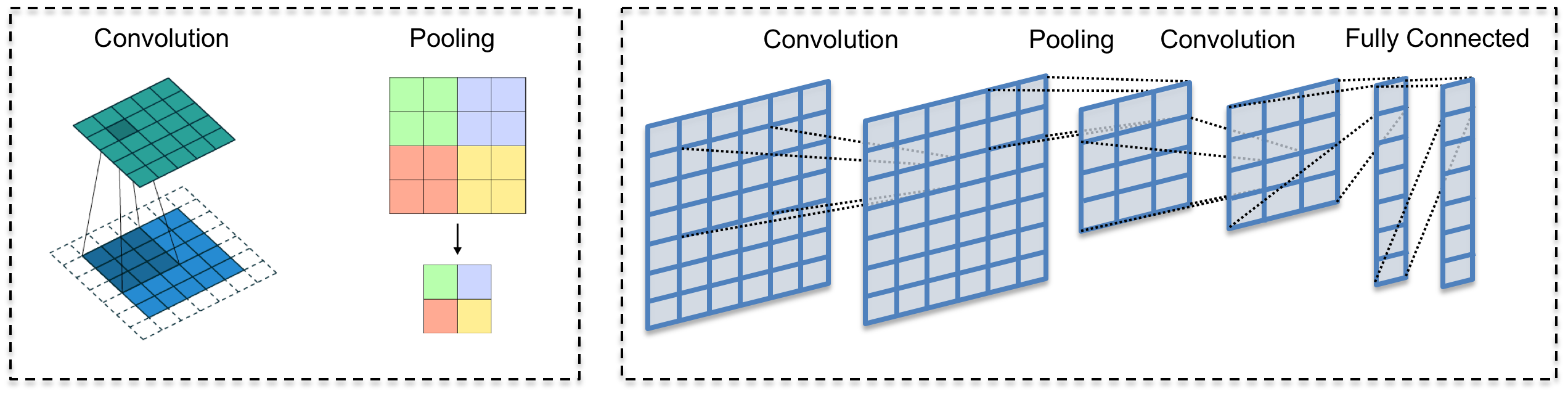}
    \caption{Convolutional layers only face a limited preceptive field and all neurons share the same weights (cf. left side of the figure; adopted from \cite{dumoulin2016guide}). Pooling layers reduce the total input size. Both are typically combined in an alternating manner to construct convolutional neural networks (CNNs). An example is shown on the right.}
    \label{fig:convpool}
\end{figure}

Another significant advance towards deep learning is the use of specialized layers. In particular, the so-called \emph{convolution} and \emph{pooling layers} enable to model locality and abstraction (cf. Fig.~\ref{fig:convpool}). The major advantage of the convolution layers is that they only consider a local neighborhood for each neuron, and that all neurons of the same layer share the same weights, which dramatically reduces the amount of {\comment parameters and therefore} memory required to store such a layer. These restrictions are identical to limiting the matrix multiplication to a matrix with circulant structure, which exactly models the operation of convolution. As the operation is generally of the form of a matrix multiplication, the gradients introduced in Section~\ref{sec:training} still apply. \emph{Pooling} is an operation that is used to reduce the scale of the input. For images, typically areas of $2\times2$ or $3\times3$ are analyzed and summarized to a single value. The average operation can again be expressed as a matrix with hard-coded weights, and gradient computation follows essentially the previous section. Non-linear operations, such as maximum or median, however, require more attention. Again, we can exploit the sub-gradient approach. During the forward pass through the net, the maximum or median can easily be determined. Once this is known, a matrix is constructed that simply selects the correct elements that would also have been selected by the non-linear methods. The transpose of the same matrix is then employed during the backward pass to determine an appropriate sub-gradient \cite{miccai:schirrmacher}. Fig.~\ref{fig:convpool} shows both operations graphically and highlights an example for a convolutional neural network (CNN). If we now compare this network with Fig.~\ref{fig:pattern_rec_pipeline}, we see that the original interpretation as only a classifier is no longer valid. Instead, the deep network now models all steps directly from the signal up to the classification stage. Hence, many authors claim that feature ``hand-crafting'' is no longer required because everything is learned by the network in a data-driven manner. 

{\commenttwo 
So far, deep learning seems quite easy. However, there are also important practical issues that all users of deep learning need to be aware of. In particular, a look at the loss over the training iterations is very important. If the loss increases quickly after the beginning, a typical problem is that the learning rate $\eta$ is set too high. This is typically referred to as \emph{exploding gradient}. Setting $\eta$ too low, however, can also result in a stagnation of the loss over iterations. In this case, we observe again {vanishing gradients}. Hence, correct choice of $\eta$ and other training hyper-parameters is crucial for successful training \cite{goodfellow2016deep}.

In addition to the training set, a validation set is used to determine over-fitting. In contrast to the training set, the validation set is never used to actually update the parameter weights. Hence, the loss of the validation set allows an estimate for the error on unseen data. During optimization, the loss on the training set will continuously fall. However, as the validation set is independent, the loss on the validation set will increase at some point in training. This is typically a good point to stop updating the model before it over-fits to the training data. 

Another common mistake is bias in training or test data. First of all, hyper-parameter tuning has to be done on validation data before actual test data is employed. In principle, test data should only be looked at once architecture, parameters, and all other factors of influence are set. Only then the test data is to be used. Otherwise, repeated testing will lead to optimistic results \cite{goodfellow2016deep} and the system's performance will be over-estimated. This is as forbidden as including the test data in the training set.
Furthermore, confounding factors may influence the classification results. If, for example, all pathological data was collected with Scanner A and all control data was collected with Scanner B, then the network may simply learn to differentiate the two scanners instead of the identifying the disease \cite{maier2009qmos}.

Due to the nature of gradient descent, training will stop once a minimum is reached. However, due to the general non-convexity of the loss function, this minimum is likely to be only a local minimum. Hence, it is advisable to perform multiple training runs with different initialization techniques in order to estimate a mean and a standard deviation for the model performance. Single training runs may be biased towards a single more or less random initialization.

Furthermore, it is very common to use typical regularization terms on parameters, as it is commonly done in other fields of medical imaging. Here, L2- and L1-norms are common choices. In addition, regularization can also be enforced by other techniques such as \emph{dropout}, \emph{weight-sharing}, and \emph{multi-task learning}. An excellent overview is given in \cite{goodfellow2016deep}.

Also note that the output of a neural network does not equal to confidence, even if they are scaled between 0 and 1 and appear like probabilities, e.g. when using the so-called \emph{softmax} function. In order to get realistic estimates of confidence other techniques have to be employed \cite{schlemper2018bayesianrecon}.
}

The last missing remark towards deep learning is the role of availability of large amounts of data and {\comment labels or} annotations that could be gathered over the internet, the immense compute power that became available by using graphics cards for general purpose computations, and, last but not least, the positive trend towards open source software that enables users world-wide to download and extend deep learning methods very quickly. All three elements were crucial to enable this extremely fast rise of deep learning.

\subsection{\comment Important Architectures in Deep Learning}
With the developments of the previous section, much progress was made towards improved signal, image, video, and audio processing, as already detailed earlier. In this introduction, we are not able to highlight all developments, because this would go well beyond the scope of this document, and there are other sources that are more suited for this purpose \cite{bengio2013representation, goodfellow2016deep, litjens2017survey}.  Instead, we will only shortly discuss some advanced network architectures that we believe had, or will have, an impact on medical image processing.

{\bf Autoencoders} use a contracting and an expanding branch to find representations of the input of a lower dimensionality \cite{vincent2008extracting}. They do not require annotations, as the network is trained to predict the original input using loss functions such as $\loss(\param)= ||\bm \classifier(\fex) - \fex||_2^2$. Variants use convolutional networks \cite{holden2015learning}, add noise to the input \cite{vincent2010stacked}, or aim at finding sparse representations \cite{huang2007unsupervised}.

{\bf Generative adversarial networks} (GANs) employ two networks to learn a representative distribution from the training data \cite{goodfellow2016nips}. A \emph{generator network} creates new images from a noise input, while a \emph{discriminator network} tries to differentiate real images from generated images. Both are trained in an alternating manner such that both gradually improve for their respective tasks. GANs are known to generate plausible and realistically looking images. {\commenttwo So-called Wasserstein GANs can reduce instability in training \cite{arjovsky2017wasserstein}.} Conditional GANs \cite{gauthier2014conditional} allow to encode states in the process such that images with desired properties can be generated. {\comment CycleGANs} \cite{zhu2017unpaired} drive this even further as they allow to convert one image from one domain to another, for example from day to night, without directly corresponding images in the training data.

{\bf Google's inception network} is an advanced and deep architecture that was applied successfully for several tasks \cite{szegedy2015going}. Its main highlight is the introduction of the so-called \emph{inception block} that essentially allows to compute convolutions and pooling operations in parallel. By repeating this block in a network, the network can select by itself in which sequence convolution and pooling layers should be combined in order to solve the task at hand effectively.

{\bf Ronneberger's U-net} is a breakthrough towards automatic image segmentation \cite{ronneberger2015u} and has been applied successfully in many tasks that require image-to-image transforms, for example, images to segmentation masks. Like the autoencoder, it consists of a contracting and an expanding branch, and it enables multi-resolution analysis. In addition, U-net features skip connections that connect the matching resolution levels of the encoder and the decoder stage. Doing so, the architecture is able to model general high-resolution multi-scale image-to-image transforms. Originally proposed in 2-D, many extensions, such as 3-D versions, exist \cite{cciccek20163d,milletari2016v}.

{\bf ResNets} have been designed to enable training of very deep networks \cite{he2016deep}. Even with the methods described earlier in this paper, networks will not benefit from more than 30 to 50 layers, as the gradient flow becomes numerically unstable in such deep networks. In order to alleviate the problem, a so-called \emph{residual block} is introduced, and layers take the form $\bm \classifier(\fex) = \fex + \bm \classifier'(\fex)$, where $\bm \classifier'(\fex)$ contains the actual network layer. Doing so has the advantage that the addition introduces a second parallel branch into the network that lets the gradient flow from end to end. ResNets also have other interesting properties, e.g., their residual blocks behave like ensembles of classifiers \cite{veit2016residual}.

{\bf Variational networks} enable the conversion of an energy minimization problem into a neural network structure \cite{kobler2017variational}. We consider this type of network as particular interesting, as many problems in traditional medical image processing are expressed as energy minimization problems. The main idea is as follows: The energy function is typically minimized by optimization programs such as gradient descent. Thus, we are able to use the gradient of the original problem to construct a so-called \emph{variational unit} that describes exactly one update step of the optimization program. Succession of such units then describe the complete variational network. Two observations are noteworthy: First, this type of framework allows to learn operators within one variational unit, such as a sparsifying transform for compressed sensing problems. Second, the variational units generally form residual blocks, and thus variational networks are always ResNets as well.

{\bf Recurrent neural networks} (RNNs) enable the processing of sequences with long term dependencies \cite{mandic2001recurrent}. Furthermore, recurrent nets introduce state variables that allow the cells to carry memory and essentially model any finite state machine. Extensions are long-short-term memory (LSTM) networks \cite{hochreiter1997long} and gated recurrent units (GRU) \cite{chung2014empirical} that can model explicit read and write memory transactions similar to a computer.

{\comment \subsection{Advanced deep learning concepts}
In addition to the above mentioned architectures, there are also useful concepts that allow building more robust and versatile networks. Again, the here listed methods are incomplete. Still, we aimed at including the most useful ones.

{\bf Data augmentation} In data augmentation, common sources of variation are explicitly added to training samples. These models of variation typically include noise, changes in contrast, and rotations and translations. In biased data, it can be used to improve the numbers of infrequent observations. In particular, the success of U-net is also related to very powerful augmentation techniques that include, for example, non-rigid deformations of input images and the desired segmentation \cite{ronneberger2015u}.} {\commenttwo In most recent literature, reports are found that also GANs are useful for data augmentation \cite{DBLP:journals/corr/abs-1803-01229}.}

{\bf Precision learning} is a strategy to include known operators into the learning process \cite{maier2018precision}. While this idea is counter-intuitive for most recognition tasks, where we want to learn the optimal representation, the approach is actually very useful for signal processing tasks in which we know \emph{a priori} that a certain operator must be present in the processing chain. Embedding the operator in the network reduces the maximal training error, reduces the number of unknowns and therefore the number of required training samples, and enables mixing of most signal processing methods with deep learning. The approach is applicable to a broad range of operators. The main requirement is that a gradient or sub-gradient must exist. 

{\bf Adversarial examples} consider the input to a neural network as a possible weak spot that could be exploited by an attacker \cite{yuan2017adversarial}. 
{\comment Generally, attacks try to find a perturbation $\bm e$ such that $\hat{f}(\bm x + \bm e)$ indicates a different class than the true $y$, while keeping the magnitude of $\bm e$ low, for example, by minimizing $||\bm e||_2^2$. Using different objective functions allows to form different types of attacks.
Attacks range from generating noise that will mislead the network, but will remain unnoticed by a human observer, to specialized patterns that will even mislead networks after printing and re-digitization \cite{brown2017adversarial}.}

{\bf Deep reinforcement learning} is a technique that allows to train an artificial agent to perform actions given inputs from an environment and expands on traditional reinforcement learning theory \cite{sutton1998reinforcement}. In this context, deep networks are often used as flexible function approximators representing value functions and/or policies \cite{silver2016mastering}. In order to enable time-series processing, sequences of environmental observations can be employed \cite{mnih2015human}.

\section{Results}

\begin{figure}
    \centering
    \includegraphics[width=\linewidth]{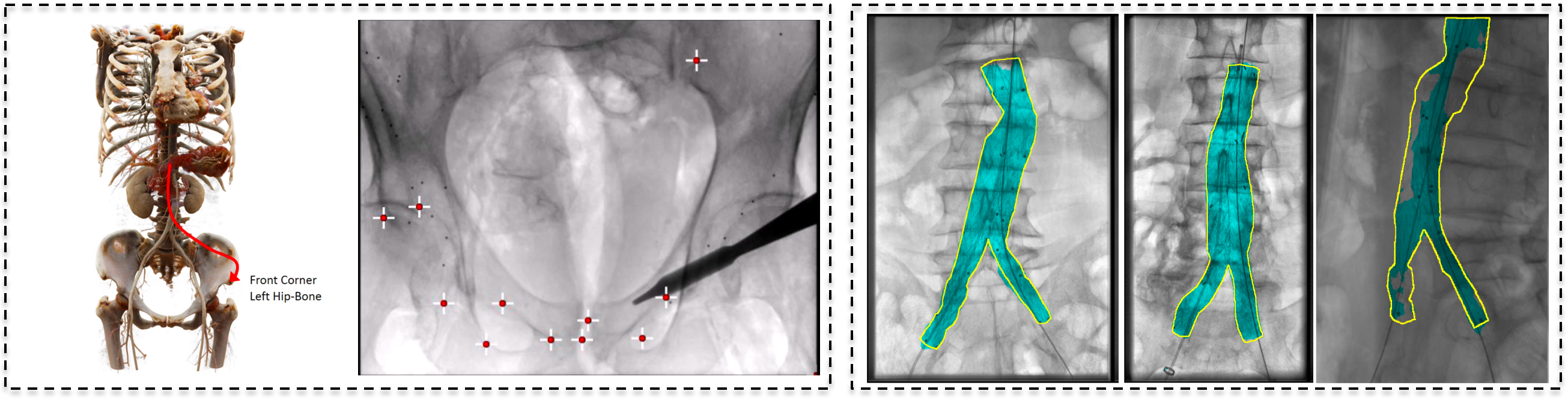}
    \caption{Deep learning excels in perceptual tasks such as detection and segmentation. The left hand side shows the artificial agent-based landmark detection after Ghesu et al. \cite{ghesu2017multi} and the X-ray transform-invariant landmark detection by Bier et al. \cite{bier2018miccai} (projection image courtesy of Dr.\,Unberath). The right hand side shows a U-net-based stent segmentation after Breininger et al. \cite{Breininger2018}. Images are reproduced with permission by the authors.}
    \label{fig:perceptual}
\end{figure}

As can be seen in the last few paragraphs, deep learning now offers a large set of new tools that are applicable to many problems in the world of medical image processing. In fact, these tools have already been widely employed. In particular, perceptual tasks are well suited for deep learning. We present some highlights that are discussed later in this section in Fig.~\ref{fig:perceptual}. On the international conference of \emph{Medical Image Computing and Computer-Assisted Intervention} (MICCAI) in 2018, approximately 70\,\% of all accepted publications were related to the topic of deep learning.  Given this fast pace of progress, we are not able to describe all relevant publications here. Hence, this overview is far from being complete. Still we want to highlight some publications that are representative for the current developments in the field. In terms of structure and organization, we follow \cite{zhou2017deep} here, but add recent developments in physical simulation and image reconstruction.

\subsection{Image detection and recognition}

Image detection and recognition deals with the problem of detecting a certain element in a medical image. In many cases, the images are volumetric. Therefore efficient parsing is a must.
A popular strategy to do so is marginal space learning \cite{zheng2014marginal}, as it is  efficient and allows to detect organs robustly. Its deep learning counter-part \cite{ghesu2016marginal} is even more efficient, as its probabilistic boosting trees are replaced using a neural network-based boosting cascade. Still, the entire volume has to be processed to detect anatomical structures reliably. \cite{ghesuDL} drives efficiency even further by replacing the search process by an artificial agent that follows anatomy to detect anatomical landmarks using deep reinforcement learning. The method is able to detect hundreds of landmarks in a complete CT volume in few seconds.

Bier et al. proposed an interesting method in which they detect anatomical landmarks in 2-D X-ray projection images \cite{bier2018miccai}. In their method, they train projection-invariant feature descriptors from 3-D annotated landmarks using a deep network.
Yet another popular method for detection are the so-called region proposal convolutional neural networks. In \cite{akselrod2016region} they are applied to robustly detect tumors in mammographic images.
 
Detection and recognition are obviously also applied in many other modalities and a great body of literature exists. Here, we only report two more applications. In histology, cell detection and classification is an important task, which is tackled by Aubreville et al. using guided spatial transformer networks \cite{aubreville} that allow refinement of the detection before the actual classification is done. The task of mitosis classification benefits from this procedure.
Convolutional neural networks are also very effective for other image classification tasks. In
 \cite{aubreville2018IJCARS} they are employed to automatically detect images containing motion artifacts in confocal laser-endoscopy images.

\subsection{Image segmentation} 

Also image segmentation greatly benefited from the recent developments in deep learning. In image segmentation, we aim to determine the outline of an organ or anatomical structure as accurately as possible. Again, approaches based on convolutional neural networks seem to dominate. Here, we only report Holger Roth's Deeporgan \cite{roth2015deeporgan}, the brain MR segmentation using CNN by Moeskops et al. \cite{moeskops2016automatic}, a fully convolutional multi-energy 3-D U-net presented by Chen et al. \cite{univis91841629}, and a U-net-based stent segmentation in X-ray projection domain by Breininger et al. \cite{Breininger2018} as representative examples. Obviously segmentation using deep convolutional networks also works in 2-D as shown by Nirschl et al. for histopathologic images \cite{nirschl2017deep}.

Middelton et al. already experimented with the fusion of neural networks and active contour models in 2004 well before the advent of deep learning \cite{middleton2004segmentation}. Yet, their approach is neither using deep nets nor end-to-end training, which would be desirable for a state-of-the-art method. Hence, revisiting traditional segmentation approaches and fusing them with deep learning in an end-to-end fashion seems a promising scope of research. Fu et al. follow a similar idea by mapping Frangi's vesselness into a neural network \cite{ArXivWeilin}. They demonstrate that they are able to adjust the convolution kernels in the first step of the algorithm towards the specific task of vessel segmentation in ophthalmic fundus imaging.

Yet another interesting class of segmentation algorithms is the use of recurrent networks for medical image segmentation. Poudel et al. demonstrate this for a recurrent fully convolutional neural network on multi-slice MRI cardiac data \cite{poudel2016recurrent}, while Andermatt et al. show effectiveness of GRUs for brain segmentation \cite{andermatt2016multi}.

\subsection{Image registration}

While the perceptual tasks of image detection and classification have been receiving a lot of attention with respect to applications of deep learning, image registration has not seen this large boost yet. However, there are several promising works found in the literature that clearly indicate that there are also a lot of opportunities.

One typical problem in point-based registration is to find good feature descriptors that allow correct identification of corresponding points. Wu et al. propose to do so using autoencoders to mine good features in an unsupervised way \cite{wu2016scalable}. Schaffert et al. drive this even further and use the registration metric itself as loss function for learning good feature representations \cite{schaffert2018metric}. Another option to solve 2-D/3-D registration problems is to estimate the 3-D pose directly from the 2-D point features \cite{miao2017convolutional}.

For full volumetric registration, examples of deep learning-based approaches are also found. The quicksilver algorithm is able to model a deformable registration and uses a patch-wise prediction directly from the image appearance \cite{yang2017quicksilver}. Another approach is to model the registration problem as a control problem that is dealt with using an agent and reinforcement learning. Liao et al. propose to do so for rigid registration predicting the next optimal movement  in order to align both volumes \cite{liao2017artificial}. This approach can also be applied to non-rigid registration using a statistical deformation model \cite{univis91731175}. In this case, the actions are movements in the vector space of the deformation model.
Obviously, agent-based approaches are also applicable for point-based registration problems. Zhong et al. demonstrate this for intra-operative brain shift using imitation learning \cite{univis91890067}.

\subsection{Computer-aided diagnosis} 

Computer-aided diagnosis is regarded as one of the most challenging problems in the field of medical image procesing. Here, we are not only acting in a supportive role quantifying evidence towards the diagnosis. Instead the diagnosis itself is to be predicted. Hence, decisions have to be done with utmost care and decisions have to be reliable.

The analysis of chest radiographs comprises a significant amount of work for radiologistic and is performed routinely. Hence, reliable support to prevent human error is highly desirable. An example to do so is given in \cite{diamant2017chest} by Diamant et al. using transfer learning techniques.

A similar workload is imposed on ophthalmologists in the reading of volumetric optical coherence tomography data. Google's Deep Mind just recently proposed to support this process in terms of referral decision support \cite{de2018clinically}.

There are many other studies found in this line, for example, automatic cancer assessment in confocal laser endoscopy in different tissues of the head and neck \cite{aubreville2017epithelialcancer}, deep learning for mammogram analysis \cite{carneiro2017deep}, and classification of 
skin cancer \cite{esteva2017dermatologist}.

\subsection{Physical simulation}
A new field of deep learning is the support of physical modelling. So far this has been exploited in the gaming industry to compute realistically appearing physics engines \cite{wu2015galileo}, or for smoke simulation \cite{chu2017data} in real-time. A first attempt to bring deep learning to bio-medical modelling was done by Meister et al. \cite{meister18:TFB}.

Based on such observations, researchers started to bring such methods into the field of medical imaging. One example to do so is the deep scatter estimation by Maier et al. \cite{maier2018deep}. Unberath et al. drive this even further to emulate the complete X-ray formation process in their DeepDRR \cite{10.1007/978-3-030-00937-3_12}. In~\cite{horger2018towards} Horger et al. demonstrate that even noise of unknown distributions can be learned, leading to an efficient generative noise model for realistic physical simulations.

Also other physical processes have been investigated using deep learning. In \cite{maier2018precision} a material decomposition using deep learning embedding prior physical operators using precision learning is proposed. Also physically less plausible interrelations are attempted. In \cite{han2017mr}, Han et al. attempt to convert MR volumes to CT volumes. Stimpel  et al. drive this even further predicting X-ray projections from MR projection images  \cite{univis91895709}. While these observations seem promising, one has to follow such endeavors with care. Schiffers et al. demonstrate that cycleGANs may create correctly appearing flourecence images from fundus images in ophthalmology \cite{schiffers2018cyclegan}. Yet, undesired effects appear, as occasionally drusen are mapped onto micro aneurysms in this process. Cohen et al. demonstrate even worse effects \cite{Cohen2018distribution}. In their study, cancers disappeared or were created during the modality-to-modality mapping. Hence, such approaches have to be handled with care. 

\subsection{Image Reconstruction}
\begin{figure}
    \centering
    \includegraphics[width=\linewidth]{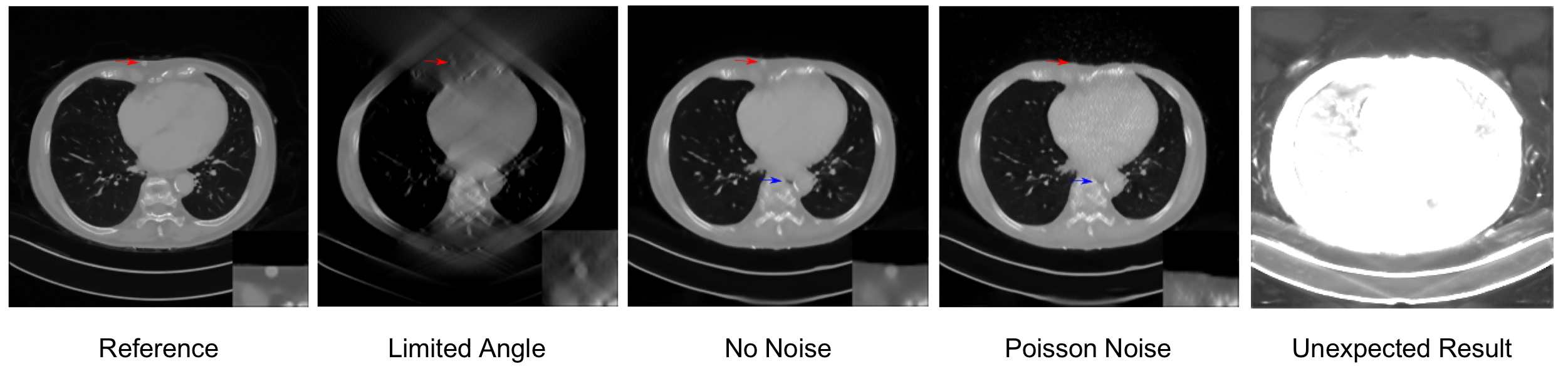}
    \caption{Results from a deep learning image-to-image reconstruction based on U-net. The reference image with a lesion embedded is shown on the left followed by the analytic reconstruction result that is used as input to U-net. U-net does an excellent job when trained and tested without noise. If unmatched noise is provided as input, an image is created that appears artifact-free, yet not just the lesion is gone, but also the chest surface is shifted by approximately 1\,cm. On the right hand side, an undesirable result is shown that emerged at some point during training of several different versions of U-net which shows organ-shaped clouds in the air in the background of the image. {\comment Note that we omitted displaying multiple versions of ``Limited Angle'' as all three inputs to the U-Nets would appear identically given the display window of the figure of [-1000, 1000] HU.}}
    \label{fig:dlrecon}
\end{figure}

Also the field of medical image reconstruction has been affected by deep learning and was just recently the topic of a special issue in the IEEE Transactions on Medical Imaging. The editorial actually gives an excellent overview on the latest developments \cite{wang2018image} that we will summarize in the next few lines.

One group of deep learning algorithms omit the actual problem of reconstruction and formulate the inverse as image-to-image transforms with different initialization techniques before processing with a neural network. Recent developments in this \emph{image-to-image reconstruction}
are summarized in \cite{mccann2017review}. Still, there is continuous progress in the field, e.g. by application of the latest network architectures \cite{zhang2018sparse} or cascading of U-nets \cite{kofler2018u}.

A recent paper by Zhu et al. proposes to learn the entire reconstruction operation only from raw data and corresponding images \cite{zhu2018image}. The basic idea is to model an autoencoder-like dimensionality reduction in raw data and reconstruction domain. Then both are linked using a non-linear correlation model. The entire model can then be converted into a single network and trained in an end-to-end manner. In the paper, they show that this is possible for 2-D MR and PET imaging and largely outperforms traditional approaches.

Learning operators completely data-driven carries the risk that undesired effects may occur \cite{huang2018considerations}, as is shown in Fig.~\ref{fig:dlrecon}. Hence integration of prior knowledge and the structure of the operators seems beneficial, as already described in the concept of precision learning in the previous section. Ye et al. embed a multi-scale transform into the encoder and decoder of a U-net-like network, which gives rise to the concept of deep convolutional framelets \cite{ye2018deep}. Using wavelets for the multi-scale transform has been successfully applied in many applications ranging from denoising \cite{kang2018deep} to sparse view computed tomography \cite{han2018framing}.

If we design a neural network inspired by iterative algorithms that minimize an energy function step by step, the concept of variational networks is useful. Doing so allows to map virtually all iterative reconstruction algorithms onto deep networks, e.g., by using a fixed number of iterations. There are several impressive works found in the literature, of which we only name the MRI reconstruction by Hammernik et al. \cite{hammernik2018learning} and the sound speed reconstruction by Vishnevskiy et al. \cite{vishnevskiy2018image} at this point. The concept can be expanded even further, as Adler et al. demonstrate by learning an entire primal-dual reconstruction \cite{adler2018learned}. 

W{\"u}rfl et al. also follow the idea of using prior operators \cite{deeplearningct, wurfl2018deep}. Their network is inspired by the classical filtered back-projection that can be retrained to better approximate limited angle geometries that typically cannot be solved by classical analytic inversion models. Interestingly, as the approach is described in an end-to-end fashion, errors in the discretization or initialization of the filtering steps are intrinsically corrected by the learning process \cite{ISBIArchiveSyben}.
They also show that their method is compatible with other approaches, such as variational networks that are able to learn an additional de-streaking sparsifying transform \cite{hammernik2017dlct}. Syben et al. drive these efforts even further and demonstrate that the concept of precision learning is able to mathematically derive a neural network structure \cite{syben2018deriving}. In their work, they demonstrate that they are able to postulate that an expensive matrix inverse is a circulant matrix and hence can be replaced by a convolution operation. Doing so leads to the derivation of a previously unknown filtering, back-projection, re-projection-style rebinning algorithm that intrinsically suffers less from resolution loss than traditional interpolation-based rebinning methods.

As noted earlier, all networks are prone to adversarial attacks. Huang et al. demonstrate this \cite{huang2018considerations} in their work, showing that already incorrect noise modelling may distort the entire image. Yet, the networks reconstruct visually pleasing results and artifacts cannot be as easily identified as in classical methods. One possible remedy is to follow the precision learning paradigm and fix as much of the network as possible, such that it can be analyzed with classical methods as demonstrated in \cite{wurfl2018deep}. Another promising approach is
Bayesian deep learning \cite{schlemper2018bayesianrecon}. Here the network output is two-fold: the reconstructed image plus a confidence map on how accurate the content of the reconstructed image was actually measured.

Obviously, deep learning also plays a role in suppression of artifacts. In~\cite{zhang2018convolutional}, Zhang et al. demonstrate this effectively for metal artifacts. As a last example, we list Bier et al. here, as they show that deep learning-based motion tracking is also feasible for motion compensated reconstruction \cite{bier2018detecting}.

\section{Discussion}
\label{S:discussion}

In this introduction, we reviewed the latest developments in deep learning {\comment for medical imaging}. In particular detection, recognition, and segmentation tasks are well solved by the deep learning algorithms. Those tasks are clearly linked to perception and there is essentially no prior knowledge present. Hence, state-of-the-art architectures from other fields, such as computer vision, can often be easily adopted to medical tasks. In order to gain better understanding of the black box, reinforcement learning and modelling of artificial agents seem well suited. 

In image registration, deep learning is not that broadly used. Yet, interesting approaches already exist that are able to either predict deformations directly from the image input, or take advantage of reinforcement learning-based techniques that model registration as on optimal control problem. Further benefits are obtained using deep networks for learning representations, which are either done in an unsupervised fashion or using the registration metric itself.

Computer-aided diagnosis is a hot topic with many recent publications address. We expect that simpler standard tasks that typically result in a high workload for medical doctors will be solved first. For more complex diagnoses, the current deep nets that immediately result in a decision are not that well suited, as it is difficult to understand the evidence. Hence, approaches are needed that link observations to evidence to construct a line of argument towards a decision. It is the strong belief of the authors that only if such evidence-based decision making is achieved, the new methodology will make a significant impact to computer-aided diagnosis.

Physical simulation can be accelerated dramatically with realistic outcomes as shown in the field of computer games and graphics. Therefore, the methods are highly relevant, in particular for interventional applications, in which real-time processing is mandatory. First approaches exist, yet there is considerable room for more new developments. In particular, precision learning and variational networks seem to be well suited for such tasks, as they provide some guarantees to prediction outcomes. Hence, we believe that there are many new developments to follow, in particular in radiation therapy and real-time interventional dose tracking.

Reconstruction based on data-driven methods yield impressive results. Yet, they may suffer from a ``new kind'' of \emph{deep learning artifacts}. In particular, the work by Huang et al. \cite{huang2018considerations} show these effects in great detail. Both precision learning as well as Bayesian approaches seem well suited to tackle the problem in the future. Yet, it is unclear how to benefit best from the data-driven methods while maintaining intuitive and safe image reading.

A great advantage of all the deep learning methods is that they are inherently compatible to each other and to many classical approaches. This fusion will spark many new developments in the future. In particular, the fusion on network-level using either the direct connection of networks or precision learning allows end-to-end training of algorithms. The only requirement for this deep fusion is that each operation in the hybrid net has a gradient or sub-gradient for the optimization. In fact, there are already efforts to design whole programming languages to be compatible with this kind of \emph{differential programming} \cite{li2018differentiable}.
With such integrated networks, multi-task learning is enabled, for example, training of networks that deliver optimal reconstruction quality and the best volumetric overlap of the resulting segmentation at the same time, as already conjectured in \cite{Wang2016DeepImaging}. This point may even be expanded to computer-aided diagnosis or patient benefit.

{\comment In general, we observe that the CNN architectures that emerge from deep learning are computationally very efficient. Networks find solutions that are on par or better than many state-of-the-art algorithms. However, their computational cost at inference time is often much lower than state-of-the-art algorithms in typical domains of medical imaging in detection, segmentation, registration, reconstruction, and physical simulation tasks. This benefit at run-time comes at high computational cost during training that can take days even on GPU clusters. Given an appropriate problem domain and training setup, we can thus exploit this effect to save run-time at the cost of additional training time.}

Deep learning is extremely data hungry. This is one of the main limitations that the field is  currently facing, and performance grows only logarithmically with the amount of data used \cite{googlepaper}. Approaches like weakly supervised training \cite{oquab2015object} will only partially be able to close this gap. Hence, one hospital or one group of researchers will not be able to gather a competitive amount of data in the near future. As such, we welcome initiatives such as the grand challenges\footnote{\url{https://grand-challenge.org}} or medical data donors\footnote{\url{http://www.medicaldatadonors.org}}, and hope that they will be successful with their mission. 

\section{Conclusion}

In this short introduction to deep learning in medical image processing we were aiming at two objectives at the same time. On the one hand, we wanted to introduce to the field of deep learning and the associated theory. On the other hand, we wanted to provide a general overview on the field and potential future applications. In particular, perceptual tasks have been studied most so far. However, with the set of tools presented here, we believe many more problems can be tackled.
So far, many problems could be solved better than the classical state-of-the-art does alone, which also sparked significant interest in the public media. Generally, safety and understanding of networks is still a large concern, but methods to deal with this are currently being developed. Hence, we believe that deep learning will probably remain an active research field for the coming years.

If you enjoyed this introduction, we recommend that you have a look at our video lecture that is available at \url{https://www.video.uni-erlangen.de/course/id/662}.

\section*{Acknowledgements}

We express our thanks to Katharina Breininger, Tobias W{\"u}rfl, and Vincent Christlein, who did a tremendous job when we created the deep learning course at the University of  Erlangen-Nuremberg. Furthermore, we would like to thank Florin Ghesu, Bastian Bier, Yixing Huang, and again Katharina Breininger for the permission to highlight their work and images in this introduction. Last but not least, we also express our gratitude to the participants of the course ``Computational Medical Imaging\footnote{\url{https://www5.cs.fau.de/lectures/sarntal-2018/}}'', who were essentially the test audience of this article during the summer school ``Ferienakademie 2018''.





\bibliographystyle{model1-num-names}
\bibliography{sample.bib}







\end{document}